\pdfoutput=1
\documentclass[letterpaper]{article} % DO NOT CHANGE THIS
\usepackage{aaai2026}  % DO NOT CHANGE THIS
\usepackage{times}  % DO NOT CHANGE THIS
\usepackage{helvet}  % DO NOT CHANGE THIS
\usepackage{courier}  % DO NOT CHANGE THIS
\usepackage[hyphens]{url}  % DO NOT CHANGE THIS
\usepackage{graphicx} % DO NOT CHANGE THIS
\usepackage{natbib}  % DO NOT CHANGE THIS AND DO NOT ADD ANY OPTIONS TO IT
\usepackage{caption} % DO NOT CHANGE THIS AND DO NOT ADD ANY OPTIONS TO IT
\usepackage{times}
\usepackage{latexsym}

\usepackage[T1]{fontenc}
\usepackage[utf8]{inputenc}

\usepackage{microtype}

\usepackage{inconsolata}

\usepackage{graphicx}
\usepackage{listings}
\usepackage{algorithm}
\usepackage{amsmath}
\usepackage{algpseudocode}
\usepackage{multirow}
\newcommand{\data}{{\fontfamily{qag}\selectfont NormHint}}
\title{"Hiding in Plain Sight": Designing Synthetic Dialog Generation for Uncovering Socially Situated Norms}

\author{Chengfei Wu, Dan Goldwasser}
\affiliations{
    Purdue University, USA\\
    \texttt{wu1491@purdue.edu}, \texttt{dgoldwas@purdue.edu}
}

\begin{document}
\maketitle

\begin{abstract}
Naturally situated conversations encapsulate the social norms inherent to their context, reflecting both the relationships between interlocutors and the underlying communicative intent. We propose a novel, multi-step framework for generating dialogues that automatically uncovers social norms from rich, context-laden interactions through a process of self-assessment and norm discovery, rather than relying on predefined norm labels. Leveraging this framework, we construct \data{}, a comprehensive synthetic dialogue dataset spanning a wide range of \textit{interlocutor attributes} (e.g., age, profession, personality), \textit{relationship types}, \textit{conversation topics}, and \textit{conversational trajectories}. \data{} is meticulously annotated with turn-level norm violation information, detailed participant descriptions, and remediation suggestions—including alternative trajectories achieved through early intervention. Our human validation and automated analysis demonstrate that our dataset captures diverse conversational topics with high naturalness and realism. We also discovered that fine-tuning a model with our norm violation data enhances its ability to detect and understand potential norm violations in conversations.
\end{abstract}

\section{Introduction}\label{sec:intro}
Humans excel at navigating complex social interactions by adapting behavior to context—what is appropriate when joking with a friend at a party may be unacceptable at a funeral. Through experience, we internalize social norms: informal rules that govern behavior in groups and societies \citep{social-norm-def}. Computational systems that interact with people must therefore reason about norms, including when and how they are violated.

Recent efforts have begun to operationalize social norms for NLP. The Linguistic Data Consortium (LDC)\footnote{\url{https://www.ldc.upenn.edu/}} has used experts to label norm adherence and violations \citep{ldc}. However, obtain authentic natural conversation can be hard and data scraped from sources such as YouTube and discussion forums rarely contains explicit violations. On the other hand, purely synthetic approaches \citep{li-etal-2023-normdial, zhan-etal-2024-renovi} that prompt LLMs to “break norms” often produce unnatural exchanges or lack the rich situational context needed for interpretation.

\begin{figure*}[!htbp]
\centering
\includegraphics[width=\textwidth]{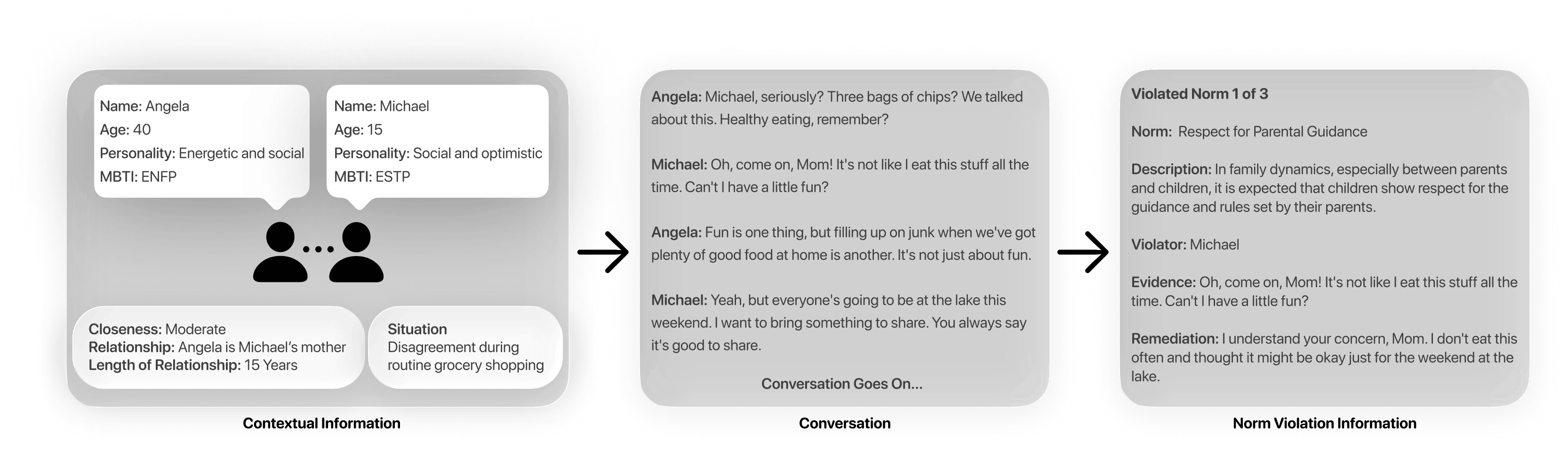}
\caption{Overview of the multi-step framework and annotation schema.}
\label{fig:teaser}
\end{figure*}

We address these limitations with a multi-step generation framework that creates diverse, context-rich dialogues first and uncovers the relevant social norms afterward. Instead of conditioning generation on predefined norms, we elicit scenarios with detailed roles, relationships, and histories, then apply a post-hoc self-assessment and norm-discovery stage to detect subtle violations and propose remediations across varied social settings. Figure \ref{fig:teaser} previews this pipeline.

Building on this framework, we introduce \data{}, a curated dataset of $1{,}743$ conversations totaling $23{,}423$ utterances, with $5{,}709$ turn-level norm violations and paired remediation suggestions. Each scenario centers on realistic conflicts or escalations and provides rich participant attributes (e.g., names, ages, personalities/MBTI \citep{mbti}, relationship closeness, and acquaintance length) spanning more than 20 relationship types. In addition, \data{} also supplies detailed situational context, turn-level labels, intent-preserving rephrases that avoid violations, and alternative trajectories obtained by intervening at the first violation with the suggested correction.

To evaluate quality and realism, we combine human evaluation with automated analysis. Automated analysis shows \data{} is up to $10\%$ more diverse than scraped situational data and outperforms other synthetic datasets by $27\%$. Human evaluation indicates that $96\%$ of scenarios are realistic given their context, and overall naturalness matches or exceeds existing resources; full details appear in Section \ref{sec:human_validation}.

In summary, this work contributes: \textbf{(1)} A novel, multi-step framework that help to discover social norms from context-rich dialogues via post-hoc self-assessment, instead of relying on predefined norms. \textbf{(2)} \data{}, a high-quality dataset with turn-level violation labels, rich context and attributes, remediation suggestions, and counterfactual continuations. \textbf{(3)} Empirical evidence that fine-tuning with our violation data improves a model’s ability to recognize potential norm violations in conversation.

\section{Related Work}\label{sec:r_work}
\subsection*{Computational Social Intelligence}

Computational social intelligence increasingly studies socio-cultural norms—implicit rules that guide acceptable behavior—and how to encode them. Work on norm identification and knowledge bases (NormKB) has used both automated and manual methods \citep{fung-etal-2023-normsage,forbes-etal-2020-social,pujari-goldwasser-2025-llm}, including uncovering nuanced, region-specific norms \citep{no-norm-left-behind}. Another line generates synthetic dialogues that either adhere to or violate norms. \citet{li-etal-2023-normdial} follow a top-down recipe: propose category-specific norms, negate them to induce violations, pair each with scenarios, and prompt ChatGPT to produce conversations—an approach that can yield contrived violations.

We instead use a bottom-up pipeline: first generate contextual-rich profiles for character pairs and plausible conflict situations; then craft conversations that naturally escalate; discover the applicable norms afterwards, similar to \citet{fung-etal-2023-normsage}. Because norms are highly context-dependent, this pipeline can also support future expansion of existing NormKBs. In addition, our dataset attaches concrete intervention suggestions to every detected violation to mitigate escalation; to our knowledge, this is the first dataset to introduce interventions.

\subsection*{Conversational Dataset}

Obtaining real conversational data is difficult due to privacy and collection costs. Pre-LLM datasets typically came from: (1) human-authored dialogues—via scraping \citep{li-etal-2017-dailydialog}, hiring actors \citep{iemocap}, or crowdsourcing \citep{aws-chat,rashkin-etal-2019-towards}—which can be short, domain-limited (e.g., empathy-primarily), and expensive; (2) TV/movie transcripts \citep{cped,meld,chen-etal-2020-mpdd}, which skew dramatic and unrealistic; and (3) social-media threads \citep{wang-etal-2013-dataset,zhang-etal-2018-conversations,ritter-etal-2011-data}, which are noisy and weakly conversational. Our approach addresses these issues by conserving annotation resources while promoting diverse, context-rich, and natural everyday dialogues.

\section{Generation Framework}\label{sec:framework}
In this section we outline our framework: (i) generate rich context for characters and situations, (ii) produce conversations with guided flows plus self-verification \citep{weng-etal-2023-large}, and (iii) discover norms and propose interventions. The full pipeline is in Algorithm~\ref{alg:framework}; templates and generation parameters appear in Appendix \ref{sec:appd_generation}.

\subsection{Character Information}\label{sec:gen_char}
To explore interpersonal dynamics, we enumerate 20+ relationship types (e.g., familial, friendship) and follow \citep{Mairesse2007}'s discovery that interaction quality relies on intimacy, duration, and personality. We prompt ChatGPT to create character pairs with traits (MBTI, dispositions, closeness, relationship duration), while constraining the relationship type and personality contrast. An example is provided in the Contextual Information section in Figure \ref{fig:teaser}.

\subsection{Situation Information}\label{sec:gen_situation}
We align scenarios with relationship type, closeness, age, etc., to ensure plausibility (e.g., chores disputes are likelier for \verb|parent-and-child| than \verb|friends|). To avoid generic themes (e.g., \textit{career}, \textit{projects}, \textit{art}), we first filter topics via N-gram analysis, then cluster with SBERT \citep{sbert} embeddings (character names removed) and keep one representative per cluster when pairwise similarity exceeds $0.75$. % This yields diverse, contextually coherent situations.

\subsection{Full Conversation}\label{sec:gen_conv}
Given the context, we generate dialogue using flow guidance (e.g., ``start cautious, escalate as boundaries are breached'') and track each participant’s evolving emotions. This mitigates overly optimistic defaults and produces more natural trajectories.

\subsection{Post Validation}\label{sec:gen_validation}
Following \citet{weng-etal-2023-large,fung-etal-2023-normsage}, the model first summarizes a conversation, then (with greedy decoding) rates alignment with the situation and flow on a 1--5 Likert scale \citep{likert-scale} plus a True/False approval. This leverages GPT-4’s strong summarization \citep{gpt-4,gpt-3-summary-perf}. Human evaluation is also done on a randomly selected subset (Section~\ref{sec:human_validation}).

\subsection{Norm Discovery and Avoidance}\label{sec:gen_ndv}
We extract instances where emerging social norms are violated in the conversation for both parties. For each violation, the model outputs a concise category, generic norm description, violator, and cited utterance, restricted to text-observable evidence. It then proposes a minimal revision preserving intent while avoiding escalation. If any violation occurs, we intervene at the first one with the revised utterance and ask the model to complete the conversation (without flow guidance), yielding an alternative outcome.

\section{{\fontfamily{qag}\selectfont NormHint}}\label{sec:human_validation}
Building on the framework in Section \ref{sec:framework}, we curated {\fontfamily{qag}\selectfont NormHint}, a corpus of 1,743 dialogues (23,423 utterances). Basic statistics appear in Table \ref{tab:stat}. We assess data quality along five axes—(i) situational likelihood, (ii) conversational naturalness and faithfulness, (iii) linguistic diversity, (iv) norm/violation discovery, and (v) intervention quality—using both human annotators and GPT-4. We then test downstream utility via norm-violation detection. All human annotation interfaces can be found in Appendix \ref{sec:appd_mturk}.

\begin{table}[!htbp]
\centering
\begin{tabular}{lr}
\hline
Dialogues & 1743 \\
Utterances & 23423 \\
Uttr. per Dialogue \textit{(Avg)} & 13 \\
Token per Dialogue \textit{(Avg)} & 226 \\
Norm Violation per Dialogue \textit{(Avg)} & 3 \\
Token per Utterance \textit{(Avg)} & 16 \\
\hline
Remediated Dialogues & 1743 \\
\hline
\end{tabular}%
\caption{Basic Statistics of Our Dataset}
\label{tab:stat}
\end{table}

\subsection{Situational Likelihood}
We ask annotators to judge whether each generated situation (given age, relationship, intimacy, etc.) is \textit{Likely} or \textit{Unlikely} (see Fig.~\ref{fig:mturk_q1}). On 100 randomly sampled situations (3 annotators each; majority vote), \textbf{96\%} were deemed likely. Inter-annotator reliability, via Randolph’s Kappa \citep{RandolphK, irr}, is \textbf{moderate} (0.53).

\subsection{Conversational Naturalness and Faithfulness}\label{sec:naturalness}
Annotators rated naturalness on a 1--5 Likert scale \citep{likert-scale} and judged faithfulness to the provided metadata/situation (Fig.~\ref{fig:mturk_q2}). On 100 examples (3 annotators each), the mean naturalness is \textbf{4.11}, and \textbf{96\%} of dialogues are faithful. To standardize our evaluation with prior studies (e.g., \citet{li-etal-2023-normdial}), we prompt GPT-4o, differs from the model that generated the conversation, using the same rubric and instruct it to produce a chain-of-thought style explanation before providing its rating, following the approach of \citet{mmhal}; GPT-4o agrees with humans \textbf{84\%} of the time. Table \ref{tab:qc_compare} compares naturalness across datasets. Notably, DailyDialogue receives lower GPT-4o naturalness (despite being human-authored), due to its ESL-teaching origin, whereas {\fontfamily{qag}\selectfont NormHint} achieves consistently high scores (GPT-4: \textbf{4.13}; Human: \textbf{4.11}).

\begin{table}[!htbp]
\centering
\resizebox{\columnwidth}{!}{%
\begin{tabular}{ll|l}
\hline
\multicolumn{1}{c}{Dataset} & \multicolumn{1}{c|}{Annotation Type} & Naturalness \\ \hline
\multicolumn{1}{c}{\multirow{1}{*}{Daily Dialogue}} & GPT-4o & 3.0 \\ \hline
\multirow{1}{*}{NormDial} & GPT-4o & 3.9 \\ \hline
\multirow{2}{*}{{\fontfamily{qag}\selectfont NormHint}} & GPT-4o & \textbf{4.13} \\
& Human & 4.11 \\ \hline
\end{tabular}%
}
\caption{\textbf{Naturalness comparison across datasets and annotation types (higher is better).} {\fontfamily{qag}\selectfont NormHint} attains the highest naturalness under both GPT-4o--based and human evaluations, outperforming Daily Dialogue and NormDial.}
\label{tab:qc_compare}
\end{table}

\subsection{Linguistic Diversity}\label{sec:anal_diverse}
We compare {\fontfamily{qag}\selectfont NormHint} to human-crafted sets (DailyDialogue \citep{li-etal-2017-dailydialog}, Friends \citep{zhou-choi-2018-exist}, Switchboard \citep{stolcke-etal-2000-dialogue}, CaSiNo \citep{chawla-etal-2021-casino}) and the LM-generated NormDial \citep{li-etal-2023-normdial} using Distinct-$n$ \citep{li-etal-2016-diversity} and the geometric mean of $n$-gram entropies for $n\in\{1,2,3\}$ \citep{majumder-etal-2021-unsupervised}. Table \ref{tab:distinct_n_cmpr} shows {\fontfamily{qag}\selectfont NormHint} is competitive with human generated datasets and substantially surpasses NormDial (e.g., +31\%, +23\%, +14\% for bi-/tri-/4-grams; +8\% entropy vs.\ NormDial), supporting our claim that it captures diverse, real-world conversational patterns.

\begin{table}[!htbp]
\centering
\resizebox{\columnwidth}{!}{%
\begin{tabular}{lllll}
\hline
\multicolumn{1}{l|}{Dataset Name} & DD-2 & DD-3 & \multicolumn{1}{l|}{DD-4} & ENTR $\uparrow$ \\ \hline
\multicolumn{5}{c}{Non-Synthetic Dataset} \\ \hline
\multicolumn{1}{l|}{Daily Dialogue} & 0.23 & 0.54 & \multicolumn{1}{l|}{0.72} & \textbf{13.84} \\
\multicolumn{1}{l|}{Friends} & 0.29 & 0.68 & \multicolumn{1}{l|}{\textbf{0.89}} & 13.58 \\
\multicolumn{1}{l|}{Switchboard} & 0.17 & 0.45 & \multicolumn{1}{l|}{0.70} & 12.83 \\
\multicolumn{1}{l|}{CaSiNo} & 0.20 & 0.48 & \multicolumn{1}{l|}{0.72} & 11.61 \\ \hline
\multicolumn{5}{c}{Synthetic Dataset} \\ \hline
\multicolumn{1}{l|}{NormDial} & 0.26 & 0.57 & \multicolumn{1}{l|}{0.77} & 12.57 \\
\multicolumn{1}{l|}{\fontfamily{qag}\selectfont NormHint} & \textbf{0.34} & \textbf{0.70} & \multicolumn{1}{l|}{0.88} & \textit{\textbf{13.54}}\\
\hline
\end{tabular}%
}
\caption{Comparison of NormHint with other dialogue dataset using Distinct-N and N-Gram Entropy}
\vspace{-1em}
\label{tab:distinct_n_cmpr}
\end{table}

\subsection{Norm and Violation Discovery Quality}
We evaluate whether identified norms apply to the characters and whether cited utterances truly violate them. On a randomly sampled subset, human evaluations show \textbf{82\%} of the instances are valid.

\subsection{Intervention Quality}
We randomly sample and test whether ChatGPT-generated revisions preserve intent while avoiding violations. Human evaluation shows \textbf{90\%} preserve the original message; \textbf{96\%} correctly remediate. Jointly, \textbf{86.7\%} both preserve intent and reduce violation risk. A GPT-4-based analysis (as in \S\ref{sec:naturalness}) shows a \textbf{$\sim$43\%} reduction in escalation. Information preservation is high (avg \textbf{4.83}/5) while escalation drops from \textbf{3.49} to \textbf{2.00} (Table \ref{tab:remediation_eval}).

\begin{table}[!htbp]
\centering
\resizebox{\columnwidth}{!}{%
\begin{tabular}{l|c|c}
\hline
Type & Escalation $\downarrow$ & Information Preservation $\uparrow$ \\ \hline
Original & 3.49 & \multirow{2}{*}{4.83} \\
Intervened & \textbf{2.00} & \\ \hline
\end{tabular}%
}
\caption{Comparison of Escalation Levels and Information Preservation in Original and Intervened Conversations}
\label{tab:remediation_eval}
\end{table}

\subsection{Effectiveness of {\fontfamily{qag}\selectfont NormHint}}
We fine-tune \verb|Llama-3.1-8b-Instruct| \citep{llama-3} with LoRA \citep{lora} on {\fontfamily{qag}\selectfont NormHint} vs.\ NormDial (harmonized format; 10\% validation; best-checkpoint selection; details in App.~\ref{sec:appd_training}). For evaluation, we follow \citet{wang-etal-2024-semeval} on Friends with emotion-causal annotations, filtering to cases where one speaker’s positive $\rightarrow$ negative transition is attributable to another’s prior utterance (509 conversations; 1,096 instances). Treating these as positive norm-violation candidates, we ask models to detect violations. Results (Table \ref{tab:violation_detection_results}) show that {\fontfamily{qag}\selectfont NormHint}-fine-tuning yields the best matches to gold positives, outperforming both the base model and NormDial fine-tuning.

\begin{table}[!htbp]
\centering
\resizebox{0.5\columnwidth}{!}{%
\begin{tabular}{l|c}
\hline
Training Data & \multicolumn{1}{l}{Accuracy} \\ \hline
None & 16.33 \\
NormDial & 15.10 \\
{\fontfamily{qag}\selectfont NormHint} & \textbf{17.52} \\ \hline
\end{tabular}%
}
\caption{Performance comparison of norm violation detection across different training data.}
\label{tab:violation_detection_results}
\end{table}

\noindent\textbf{Summary.} {\fontfamily{qag}\selectfont NormHint} exhibits high situational plausibility, strong naturalness/faithfulness, effective and minimally escalatory interventions, and robust linguistic diversity. Crucially, its synthetic, context-aware annotations improve downstream norm-violation detection, underscoring the value of careful curation over naïve synthetic generation.

\section{Conclusion}\label{sec:conclusion}
We presented a novel multi-step generation framework that uncovers social norms directly from context-rich dialogues, rather than relying on predefined norm categories. Our approach generates natural conversations imbued with detailed contextual information, which are then analyzed to identify norm violations and suggest appropriate remediation strategies. Extensive human and automated evaluations confirm that {\fontfamily{qag}\selectfont NormHint} not primarily captures a wide range of social contexts and conversational trajectories with high naturalness but also enhances model performance; our experiments demonstrate that fine-tuning with {\fontfamily{qag}\selectfont NormHint} can improve model’s ability to detect potential norm violations.

\section{Acknowledgment}
The work was supported by NSF CAREER award IIS2048001 and the DARPA CCU program. Contents do not necessarily represent the official views of, nor an endorsement by, DARPA, or the US Government.

\bibliography{anthology,custom}
% \bibliography{custom}

\newpage

\appendix

\section*{Limitations}
The scope of this study was primarily confined to the examination of conversations with conflict in the English language. This focus inherently imposes a limitation on our exploration, as it restricts our understanding to the norms and nuances prevalent within English-speaking societies. Consequently, the potential diversity and richness of conversational norms in non-English speaking cultures remain unexplored, thereby creating a gap in our comprehensive understanding of global conversational norms.

Furthermore, our reliance on crowdsource workers from the United States, Canada, and the United Kingdom for the human validation process introduces another layer of limitation. This geographical constraint could potentially skew our findings, as the perspectives and interpretations of these workers are inevitably influenced by their cultural and societal backgrounds. The absence of input from crowdsource workers from other regions of the world might lead to a less than ideal norm discovery process, as it overlooks the diversity and complexity of global conversational norms.

In essence, while our study provides insights into the norms of nagative conversations in English-speaking societies, it is still limited based on our scopes. Future research should aim to incorporate a more diverse range of languages and cultural perspectives to achieve a more holistic and inclusive understanding of conversational norms.

\section*{Ethics Statement}
For human annotations, we paid $\$0.65$ for each annotation that estimate completion time is around 2 minutes and $\$0.45$ for annotations that estimate completion time is around 1 and half minute. This yields an hourly wage of $\$18$ and $\$19.5$ respectively, which is well above the minimum hourly wage of $\$12.9$ set by the US Federal Government for 2024\footnote{\url{https://www.govinfo.gov/content/pkg/FR-2023-09-28/pdf/2023-21114.pdf}}.

\section{Generation Template and Parameter Used}\label{sec:appd_generation}
\subsection{Template for Character Pair Generation} \label{sec:apdx_gen_char}
\begin{lstlisting}
Imagine {num_pairs} of participants for conversations, with the following requirements.

**Requirements**:
1. List their name and age first.
2. Assume they have {personal_desc} personalities, describe their personality separately in two sentences.
3. Personality should not include their hobby, it should be generic but with details. DO NOT mention each other's name, describe like they don't know each other.
4. Use the personality to come up with their MBTI. Also include a one-sentence generic explanation for that MBTI type.
5. Based on their relationship, describe how close they are using terms like "extremely close", "very close", "moderately close", "slightly close", "not close at all". They don't have to be close, but closeness must relate to the relationship given. For example, if they are siblings, they are probably have a close relationship. If they are strangers, they must not be close.
6. Describe how did they meet in a sentences with details. They don't have to know each other, it can be the first time they met. However, the description must relate to the relationship given. If they know for life, just put "since birth".
7. Describe how long have they known each other with a time. If this is the first time they met, just put "first time".
8. Generate with plain text and strictly follow the output format. If more than one pair is generated, separate each pairs by "====".

**Restrictions**:
Everything generated MUST align with their relationship of {relation_desc}.
They MUST have {personal_desc} personalities.
Each pair MUST be unique from each other.
Generate exactly {num_pairs} pairs.

**Output format**:
Name:
Age:
Personality:
MBTI:

Name:
Age:
Personality:
MBTI:

How did they meet:
How long have they known each other:
Closeness:
====
\end{lstlisting}

\subsection{Template for Situation Generation} \label{sec:apdx_gen_situation}
\begin{lstlisting}
Using the information provided below, imagine what are some scenarios where {person_1_name} or {person_2_name} will start a conversation with each other?

Name: {person_1_name}
Age: {person_1_age}
Personality: {person_1_personality}
MBTI: {person_1_mbti} {person_1_mbti_desc}

Name: {person_2_name}
Age: {person_2_age}
Personality: {person_2_personality}
MBTI: {person_2_mbti} {person_2_mbti_desc}

Closeness: {closeness}

How they know each other: {how_they_know}
How long do they know each other: {how_long_they_know}
Their relationship: {relationship}

**Restrictions**:
Avoid scenarios including: projects, discovery, social gathering, art, poem, trips, family gathering, career plans, future plans.

**Requirements**:
1. Each scenario must be common, day to day, and non-generic that is likely to happen between {relationship} at their age.
2. These scenarios should be more unique to their relationship. I.e., the same scenario is not likely to happen to other relationships.
3. These scenarios must be conditioned on their closeness. I.e., the scenarios should be more likely to happen between people who are {closeness}.
4. List as much different scenarios as possible but do not exceed total of five. Keep these scenarios to be as distinguishable and diverse as possible.
5. Each scenario should be one to three sentences long with details to make them not generic. It should only include the scenario.
6. Make sure these situation will likely to lead to a conflict that will result in an awkward, unpleasant or other negative ending.
7. Do not generate repeated or similar scenarios that have been generated previously.
8. Generate without Markdown syntax and address each one with their name. List them one by one with numbering.
\end{lstlisting}

\subsection{Template for Conversation Generation} \label{sec:apdx_gen_conv}
\begin{lstlisting}
Imagine a 5-10 turns conversation between {person_1_name} and {person_2_name} with the following information, requirements and restrictions.

Name: {person_1_name}
Age: {person_1_age}
Personality: {person_1_personality}
MBTI: {person_1_mbti} {person_1_mbti_desc}

Name: {person_2_name}
Age: {person_2_age}
Personality: {person_2_personality}
MBTI: {person_2_mbti} {person_2_mbti_desc}

Closeness: {closeness}

How they know each other: {how_they_know}
How long do they know each other: {how_long_they_know}
Relationship: {relationship}

Situation: {situation}

**Requirements**:
1. The conversation should feel natural and real to people.
2. Since participants don't often articulate their thoughts exactly, the conversation should have utterances where their true message is hidden between what they said.
3. The conversation that reflects the individuals' unique traits, such as personality, closeness, age difference, and relationship dynamics, ensuring it aligns with the provided information and avoiding generic dialogue.
4. The conversation should {flow}
5. Conversation does not need to be peaceful, there can be arguments, conflict or even curse word.
6. Adjust level of respectfulness with each one's emotional state.
7. DO NOT repeat or use similar words that is used to describe each character.
8. Avoid repeating what has been said previously in the conversation.
9. Use casual words that usually appear in a conversation.
10. Format should be "Name (Emotion): Utterance". Do not include anything else. Do not use Markdown nor double quotes for the utterances.
11. Emotion should fall into the space of Plutchik's wheel of emotion.

**Restrictions**:
Avoid using the following or similar terms in the conversation
- Trying to help
- I'm doing my best
- Don't come crying to me
- Agree to disagree
- The way I am
\end{lstlisting}

\subsection{Template for Summarize the Conversation} \label{sec:apdx_summ}
\begin{lstlisting}
{conversation}

Summarize the above conversation in 4-5 sentences. It should capture the situation and the flow of the conversation. It should also indicate the outcome of the conversation (i.e., If it ended positively or negatively).
\end{lstlisting}

\subsection{Template for Self-Verification} \label{sec:apdx_self_verify}
\begin{lstlisting}
Use the situation, conversation flow and summary given, answer the following tasks. Strictly follow the output format.

**Tasks with instruction**:
- Does summary align with both the situation and the flow? Respond with only Yes or No.
- On a scale of 1-5, rate the alignment between situation and summary. 1 being the summary is not describing the situation and 5 being the summary is describing the situation. Respond with only a number.
- On a scale of 1-5, rate the alignment between conversation flow and summary. 1 being the summary is not reflecting the flow provided and 5 being the summary is reflecting the flow. Respond with only a number.

Situation: {situation}
Conversation Flow: {flow}

Summary:
{summary}

**Output format**:
Situation: [1-5]
Flow: [1-5]
Overall Alignment: [Yes/No]
\end{lstlisting}

\subsection{Template for Norm Violation Discovery and Remediation Suggestion} \label{sec:apdx_norm_ident}
\begin{lstlisting}
The following is a conversation between {relationship}. Explain why this conversation did not go well partially or entirely. What are some of the norms or rules that are being violated in the conversation?

**Information about the participants**:
Participant 1: {person_1_name}, Age {person_1_age}
Participant 2: {person_2_name}, Age {person_2_age}
Relationship: {relationship}
Closeness: {closeness}
How they know each other: {how_they_know}
How long do they know each other: {how_long_they_know}

**Requirements**:
1. Norms or rules should be generic meaning it can be applied to different scenarios. However, norms listed must be applicable to the participants given their relationship, closeness, and other information provided.
2. Descriptions should describe the norm in a general way. Include information like what is the expected behavior. Do not mention anything from the conversation nor the information provided.
3. Analyze norm violations for both participants.
4. List only one violator for each norm. If both participants violated the same norm, list them separately.
5. Evidence should be the utterance where the violation happened. The utterance must from the violator. Only include the utterance without name and emotion.
6. Only list the norms that can be observed from the text transcript. Do not list norms that require further evidence from audio and video. An example to avoid is actively listening, as you can't observe that from the text.
7. Only list norms that significantly caused this conversation to go awry.
8. Suggest a way to make least amount of changes to the original utterance that convey the same message
 and intention without break the norm violated. The suggestion can not be generic. It must align with the conversation until that turn and the speaker. Do not repect anything said previously.
9. Do not list similar norms. Make sure listed norms are common and important.
10. Do not use Markdown, and follow the output format provided. Do not generate anything else.

**Output format**:
Norm:
Description:
Violator:
Evidence:
Suggestion:

**Conversation**:
{conversation}
\end{lstlisting}

\subsection{Template for Naturalness} \label{sec:apdx_naturalness}
\begin{lstlisting}
Instruction:
Please act as an objective judge and evaluate the naturalness of the conversation given. Read the conversation between the two participants. Pay attention to their dialogue, tone, the flow of conversation, as well as whether it resembles a typical conversation in the given context.


To evaluate the conversation, first, begin your evaluation by providing a short explanation of how likely the conversation is going to happen as well as an explanation of the naturalness. After providing your explanation, you must rate the conversation by choosing from the following options:
- Rating 5: The conversation sounds entirely natural and realistic.
- Rating 4: The conversation sounds reasonably natural and flows well.
- Rating 3: The conversation is neither particularly natural nor unnatural.
- Rating 2: The conversation lacks some naturalness but is not entirely unrealistic.
- Rating 1: The conversation sounds forced, awkward, or unrealistic.

Background Information:
Note: If relationship is unknown, try to infer it from the conversation given below.
Relationship: {relationship}

Conversation:
{conv}
\end{lstlisting}

\subsection{Template for Escalation} \label{sec:apdx_escalation}
\begin{lstlisting}
Objective: Measure the Level of Escalation in a Conversation

Instructions:
1. Start by briefly summarizing what happened in the conversation. Focus on the key turning points and the overall tone.
2. Choose a rating based on how much the conversation escalated. Use the following scale:
- Rating 5: The conversation has escalated to a high degree of conflict. (Very hostile)
- Rating 4: The conversation has escalated to a moderate degree of conflict.
- Rating 3: The conversation has escalated but to a low degree of conflict.
- Rating 2: The conversation has not escalated, but there is potential for conflict.
- Rating 1: The conversation has not escalated, and there is no potential for conflict. (No hostility)

Conversation:
{conv}
\end{lstlisting}

\subsection{Template for Intervention Conversation Quality} \label{sec:apdx_ic_qc}
\begin{lstlisting}
Objective: Compare the Overall Conversation Quality Between Two Conversations

Instructions:
1. Start by briefly summarizing what happened in each conversation. Focus on the key points and the overall tone of each conversation.
2. Compare the two conversations based on the following criteria:
  - Are the characters in both conversations conveying similar messages?
  - Is the core complaint or issue addressed in both conversations?
  - Are both conversations about the same situation or topic?
3. Choose a rating based on the overall quality of the conversations. Use the following scale:
- Rating 5: The conversations are highly similar in message, address the same core complaint, and are focused on the same situation.
- Rating 4: The conversations are mostly similar in message, address mostly the same core complaint, and are largely focused on the same situation.
- Rating 3: The conversations share some similarities in message, address somewhat the same core complaint, and are generally about the same situation.
- Rating 2: The conversations have few similarities in message, address barely similar core complaint, and are loosely related to the same situation.
- Rating 1: The conversations are not similar in message, address different core complaint, and are about different situations.

Conversation 1:
{conv_1}

Conversation 2:
{conv_2}
\end{lstlisting}

\section{Annotation UI}\label{sec:appd_mturk}
\begin{figure*}[!htbp]
    \centering
    \includegraphics[width=399pt]{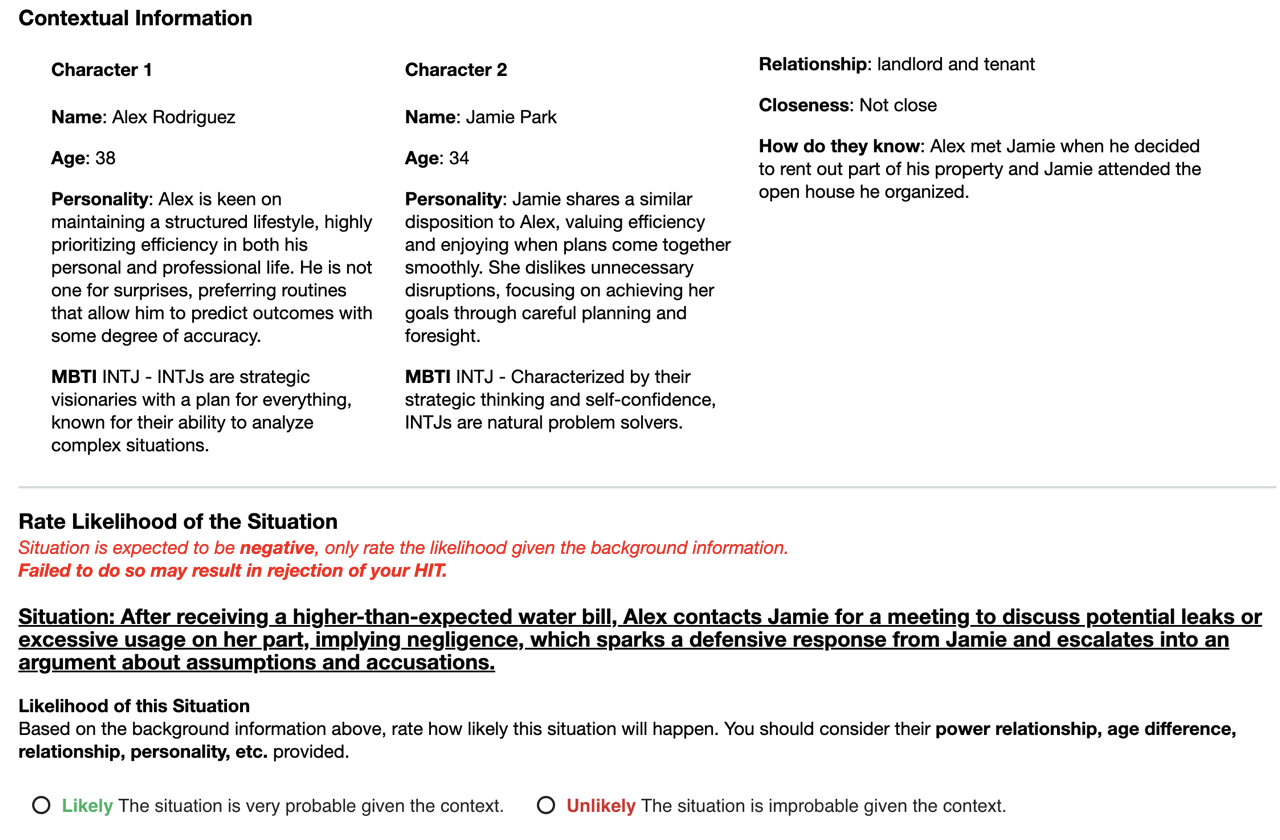}
    \caption{UI for annotator to judge likelihood of the situation}
    \label{fig:mturk_q1}
\end{figure*}

\begin{figure*}[!htbp]
    \centering
    \includegraphics[width=399pt]{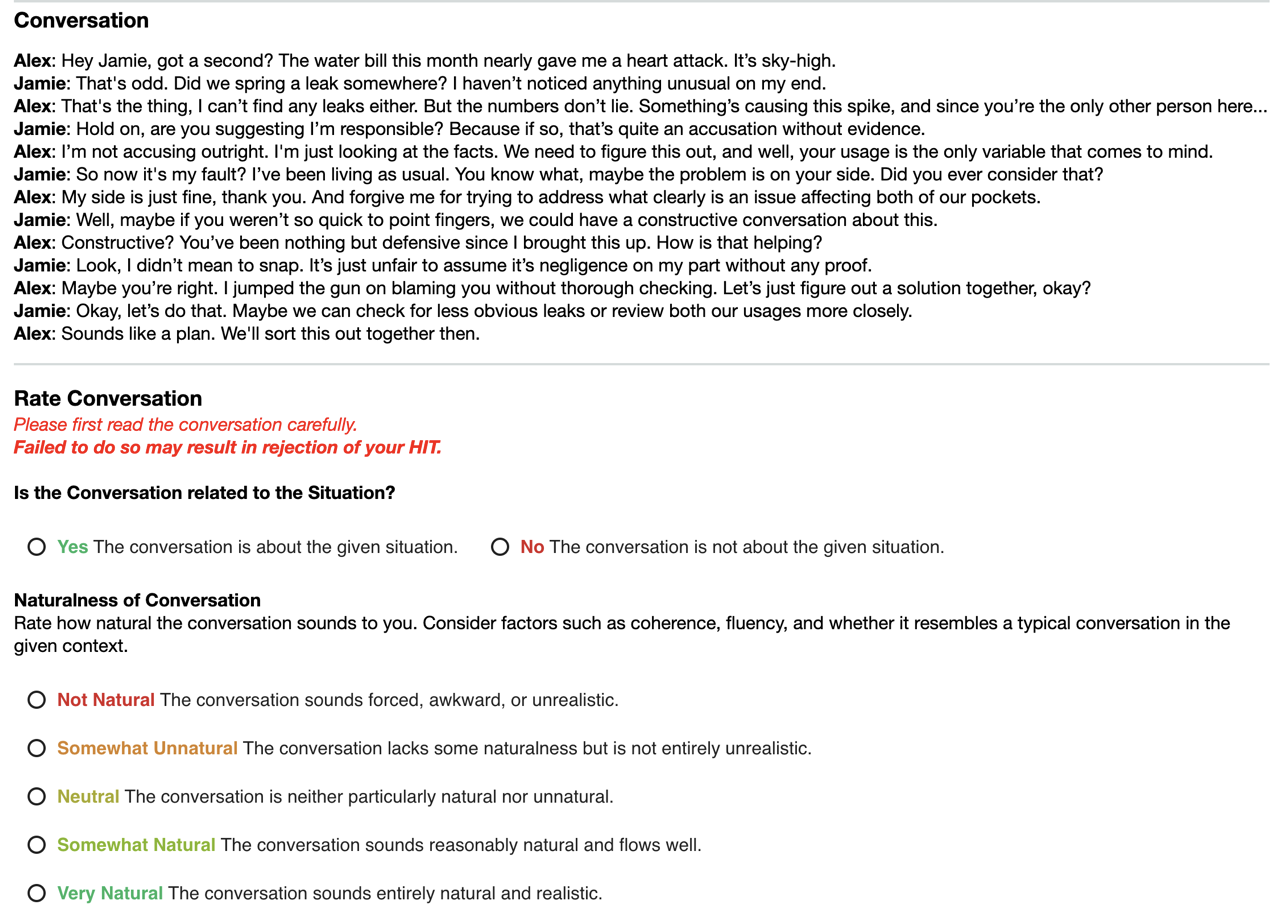}
    \caption{UI for annotator to rate naturalness of the conversation (Below) and whether the conversation aligns with the situation (Above)}
    \label{fig:mturk_q2}
\end{figure*}

\begin{figure*}[!htbp]
    \centering
    \includegraphics[width=399pt]{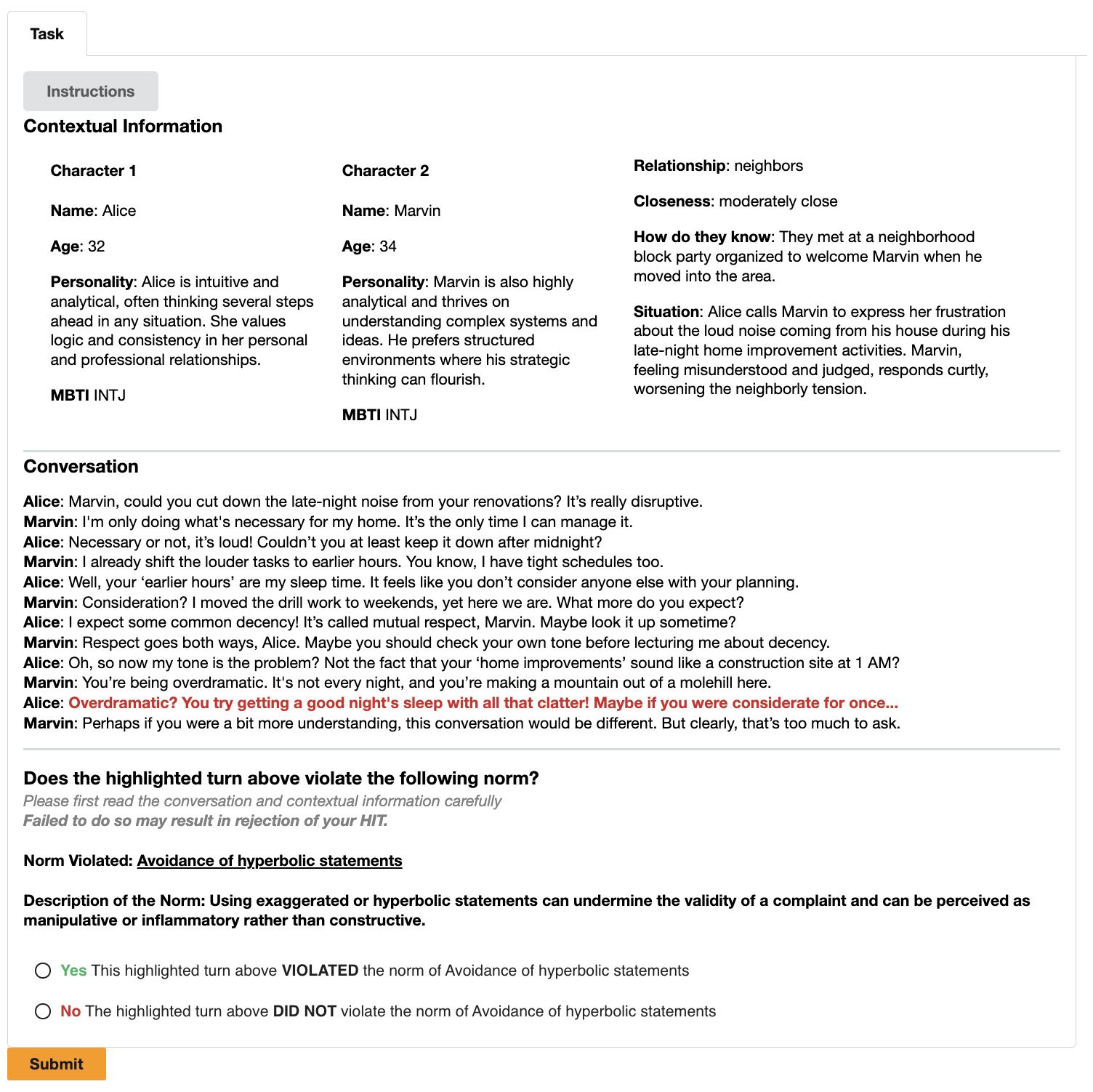}
    \caption{UI for annotator to judge the norm violation}
    \label{fig:mturk_q4}
\end{figure*}

\begin{figure*}[!htbp]
    \centering
    \includegraphics[width=399pt]{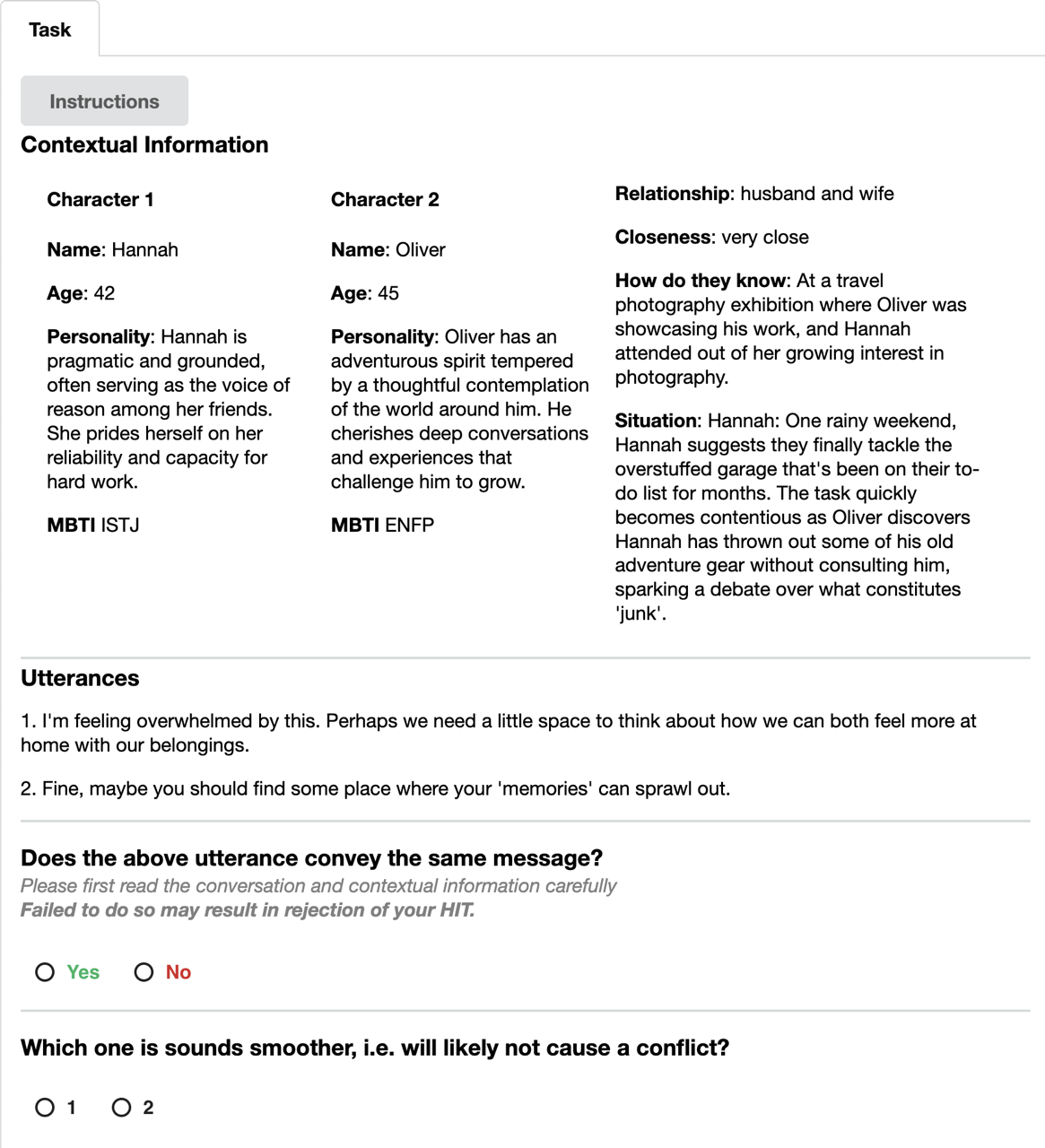}
    \caption{UI for annotator to rate remediation}
    \label{fig:mturk_q5}
\end{figure*}

\section{Correlation between Norm and Relationship}\label{sec:anal_ci}
In our study of how relationships influence the nature of norm violations, we systematically categorized conversations based on different relationship types. For each category, we identified tri-grams for norm violations and noted those appearing more than five times.

Upon analyzing each relationship type, we discovered distinct patterns in norm violations that highlight the unique dynamics within each type of relationship. 

In \verb|parent-child| relationships, the norms most frequently violated are \textit{maintaining supportive tone} and \textit{respecting personal space}. These findings underline the often hierarchical nature of this relationship, where parents may adopt a more directive tone and impose boundaries, which can lead to emotional and spatial conflicts.

In \verb|sibling| relationships, the most common norm violations include \textit{avoiding accusatory language} and \textit{respecting personal property}. This suggests that siblings, who typically share a more egalitarian and competitive dynamic, are prone to engaging in accusatory exchanges and conflicts over shared or personal items.

When examining romantic relationships, such as \verb|husband-wife| and \verb|boyfriend-girlfriend|, the prevalent norm violations shift to \textit{acknowledging partner's feelings} and \textit{addressing concerns directly}. This finding highlights the expectation for partners to be emotionally supportive and communicative, reflecting the close, intimate nature of such relationships.

In \verb|colleague| relationships, the pattern of norm violations is distinct yet again, with \textit{maintaining a professional tone} and \textit{avoiding personal attacks} being the most common. This is indicative of the professional and often formal environment of the workplace, where maintaining decorum and avoiding personal conflicts is crucial.

These observations reveal that relationship contexts significantly shape the types of norms likely to be violated, capturing the nuanced dynamics specific to each type of interpersonal interaction.

\section{Training Details}\label{sec:appd_training}

\subsection{Training Arguments}

\begin{lstlisting}
==Base Model==
unsloth/Meta-Llama-3.1-8b-Instruct-bnb-4bit

==Lora Args==
r=16
target_modules=[
    "q_proj", "k_proj", "v_proj", "o_proj",
    "gate_proj", "up_proj", "down_proj"
]
lora_alpha=16
lora_dropout=0
bias="none"
use_gradient_checkpointing="unsloth"
random_state=3407
use_rslora=False
loftq_config=None

==Training Args==
per_device_eval_batch_size=32
per_device_train_batch_size=8
gradient_accumulation_steps=2
learning_rate=2e-4
lr_scheduler_type="linear"
warmup_steps=10
num_train_epochs=4
bf16=True
optim="adamw_8bit"
weight_decay=0.01
max_grad_norm=0.3
save_strategy="best"
eval_steps=5,
eval_strategy="steps"
seed=42
greater_is_better=False
metric_for_best_model="eval_loss"
\end{lstlisting}

\subsection{Training Prompt Template}

\begin{lstlisting}
**Instruction:**
Analyze the given conversation for any violations of social norms by either participant. Identify each violation, specifying the social norm that was violated, the violator, and the turn in which the violation occurred. Additionally, suggest a more appropriate way for the violator to express themselves to avoid escalating the situation.

If no social norms were violated, respond with: `No clear violation found.` Otherwise, format your response as follows:

**Response Format:**
1. Social Norm Violated: [Brief explanation of the norm and how it was violated]
Detailed Explanation: [Provide a detailed explanation of the violation]
Violator: [Name of the person who violated the norm]
Violated Turn: [Utterance where the violation occurred]
Suggestion: [How they could have expressed themselves more appropriately]
...

**Conversation:**
{conv}"""
\end{lstlisting}

\begin{figure*}[!htbp]
    \begin{minipage}{0.98\textwidth}
        \begin{algorithm}[H]
            \caption{Conversation Generation Pipeline}\label{alg:framework}
            \begin{algorithmic}[1]
            \State \textbf{Input}: Seed relationships $R_{pool}$
            \State Initialize conversation list: $Conv \gets []$
            \State Initialize participant profiles list: $P \gets []$
            \State Define prompts: $Prompts \gets \{Pt_{pair}, Pt_{sit}, Pt_{conv}, Pt_{qc}, \dots\}$
            
            \State \textbf{// Step 1: Generate character profiles for each relationship}
            \For{each relationship $r \in R_{pool}$}
                \State Generate character profile using prompt $Pt_{pair}$:
                \State \hskip1em $P \gets P \cup OpenAI(r, Pt_{pair})$
            \EndFor
            
            \State \textbf{// Step 2: Generate potential situations for each character profile}
            \For{each profile $P_i \in P$}
                \State Generate situations using prompt $Pt_{sit}$:
                \State \hskip1em $P_i[\text{situations}] \gets OpenAI(P_i, Pt_{sit})$
            \EndFor
            
            \State \textbf{// Step 3: Filter out similar situations within the same relationship}
            \State $P_{filtered} \gets Filter(P, R_{pool})$
            
            \State \textbf{// Step 4: Generate conversations and perform quality check}
            \For{each profile $P_i \in P_{filtered}$}
                \For{each situation $s \in P_i[\text{situations}]$}
                    \State Generate conversation using prompt $Pt_{conv}$:
                    \State \hskip1em $conv \gets OpenAI(P_i, s, Pt_{conv})$
                    \State Verify conversation quality using prompt $Pt_{qc}$:
                    \State \hskip1em $isValid \gets OpenAI(conv, P_i, Pt_{qc})$
                    \If{$isValid$}
                        \State Add valid conversation to $Conv$:
                        \State \hskip1em $Conv.add(conv, P_i, s)$
                    \EndIf
                \EndFor
            \EndFor
            \end{algorithmic}
        \end{algorithm}
    \end{minipage}
    \caption{Overview of the entire pipeline from a pool of seed relationships to conversation.}
    \label{fig:framework_algo}
\end{figure*}

\end{document}